\documentclass{article}

\usepackage[final]{nips_2018}

\usepackage[utf8]{inputenc} 
\usepackage[T1]{fontenc}    
\usepackage{url}            
\usepackage{booktabs}       
\usepackage{amsfonts}       
\usepackage{nicefrac}       
\usepackage{microtype}      
\usepackage{amsmath}
\usepackage{amssymb}
\usepackage{svg}
\usepackage{graphicx,import}

\title{
Image-Level Attentional Context Modeling \\ Using Nested-Graph Neural Networks}

\author{
  Guillaume Jaume$^{1,2}$ \quad Behzad Bozorgtabar$^2$ \quad Haz{\i}m Kemal Ekenel$^{2,3}$ \\ \textbf{Jean-Philippe Thiran$^2$} \quad \textbf{Maria Gabrani$^1$}\\
  \texttt{gja@zurich.ibm.com} \quad \texttt{behzad.bozorgtabar@epfl.ch} \quad \texttt{ekenel@itu.edu.tr} \\
  \texttt{jean-philippe.thiran@epfl.ch} \quad \texttt{mga@zurich.ibm.com} \\
  $^1$IBM Research, Zurich, Switzerland \\
  $^2$\'{E}cole Polytechnique F\'{e}d\'{e}rale de Lausanne, Switzerland \\
  $^3$Istanbul Technical University, Turkey
}

\begin{document}

\maketitle

\vspace{-1mm}
\begin{abstract}
We introduce a new scene graph generation method called image-level attentional context modeling (ILAC). Our model includes an attentional graph network that effectively propagates contextual information across the graph using image-level features. Whereas previous works use an object-centric context, we build an image-level context agent to encode the scene properties. 
The proposed method comprises a single-stream network that iteratively refines the scene graph with a nested graph neural network.
%
%
We demonstrate that our approach achieves competitive performance with the state-of-the-art for scene graph generation on the Visual Genome dataset, while requiring fewer parameters than other methods. We also show that ILAC can improve regular object detectors by incorporating relational image-level information.
\end{abstract}

%

\section{Introduction} \label{sec:introduction}
Scene graph generation aims at localizing all pairs of interacting objects in an input image and identifying relationships between them in the form of graph representations, where the nodes represent objects, e.g.~\textit{a man}, \textit{a horse}, and the edges represent the pairwise interaction between objects, e.g.~man \textit{feeding} horse. In recent years, significant progress has been made in scene graph generation \cite{Xu2017, Zellers2017, Li2018, Newell2017, Li2017a}. 
One approach is to integrate external knowledge via linguistic priors \cite{Li2017a, Lu2016, Yu2017, Zhang2017a, Zhuang2017} or knowledge graphs \cite{Chen2018}. Other works have focused on building robust visual features \cite{Goutsu2018, Newell2017, Yang2018b}. However, these methods do not fully explore the potential of feature-level message sharing in order to achieve visual relationship recognition.
Motivated by recent advances in modeling neural message passing \cite{Battaglia2018, Gilmer2017, Santoro2017, Raposo2017, Online2009, Kipf2018}, we aim to extract the relational component of scene graphs by inducing a relational inductive bias in the architecture. From isolated object detection as well as visual and spatial features, we refine the object detection and infer the predicates between them. Graph Networks~\cite{Battaglia2018} introduced the idea of a global attribute that encodes shared information across the graph. Inspired by this work, we propose to use as a global attribute, an image-based context agent, that encodes image-level properties. The context aggregates information from all the nodes and edges through an attentional mechanism to retain only the relevant information.

Previous works have shown interest in relational networks for scene graph generation \cite{Xu2017, Yang2018a, Dai2017, Li2017, Li2017a, Li2018}. For instance, Xu et al.~\cite{Xu2017} proposed an iterative algorithm to pass messages between the nodes and the edges through a primal-dual graph. However, in their approach, a node can pass messages only to its neighboring edges. 
Messages between objects must pass through an edge connecting the two objects. Similarly, messages between edges must pass through objects. The information flow is therefore constrained by these node-to-edge and edge-to-node messages. In contrast, our image-level context agent allows fast information propagation across the graph. 
Recently, Li et al.~\cite{Li2018} proposed an efficient method to tackle the problem of the quadratic combinations of possible relationships with a relational proposal network. Interactions between objects, predicates and subjects are then modeled with an attention mechanism.
The importance of global context was shown by Zellers et al.~\cite{Zellers2017}, who transformed the graph topology of the problem into a sequence by stacking objects and passing them through a bi-LSTM that encodes object-based contexts. First, the context is used to infer the objects and then the predicates. The context is therefore object-centric. Unlike the previous works that benefit from object-centric context, we opt for an efficient image-level attentional context that shares information across the entire graph. Our contributions are:
\begin{enumerate}
    \item  We demonstrate the importance of image-level context for object and predicate classification. 
    \item We propose an iterative graph-propagation network to encode different semantic levels of scene understanding. The network jointly refines the object detection process by modeling its context while predicting relationships between object pairs.
    \item We propose an efficient approach that uses low-dimensional  
    edge/node representations and an image-level context. Our method achieves comparable performance with the state-of-the-art, while using less parameters.
\end{enumerate}
%

\section{The impact of context} \label{sec:context}

Previous works have shown that objects are conditionally dependent on their context \cite{Chen2017, Marino2017, Chen2018}. For instance, knowing that a \textit{tree} is in the scene raises our expectation about seeing another \textit{tree}. Behind this intuition, we can draw a homogeneity principle stating that the objects appearing in a scene tend to be compatible with their environment. To validate this idea, we studied the information gain when we partially reveal the context of an object. We computed the empirical object distribution $\Tilde{p}(o)$, derived with frequency counts from the Visual Genome training data (Xu et al. \cite{Xu2017} cleansed version). We then computed the object distribution conditioned by one randomly chosen object from the same image $I$, denoted by $\Tilde{p}(o|c_i)$, where the context $c=\{c_1, c_i,.., c_n\}$ represents the objects in $I$. Due to the limited dataset size, obtaining a meaningful empirical distribution for $\Tilde{p}(o|c)$ is not feasible. We assume that the lessons learned from studying $\Tilde{p}(o|c=c_i)$ can be generalized to $\Tilde{p}(o|c)$. To determine whether this reasoning applies to relationships, we computed the predicate distribution with and without conditioning. From the empirical distributions, we computed the entropy, conditional entropy, and the maximum entropy for the objects and the predicates (see Table~\ref{tab:entropy}). 

\begin{table}[h]
  \caption{Empirical entropy study of the Visual Genome dataset.}
  \label{tab:entropy}
  \centering
  \begin{tabular}{lccc}
  \toprule
    Category     &   $H_{max}$ & $H(x)$ & $H(x|c_i)$ \\
    \midrule
    Objects &   $5.01$  & $4.61$ & $4.08$     \\
    Predicates & $3.91$ & $2.44$ & $2.37$     \\
    \bottomrule
  \end{tabular}
\end{table}

By exposing an object to its context, we observe a significant decrease of its entropy, implying that conditioning can help object detection. To a lesser extent, the same behavior is observed for predicates.
Predicate classification is also a strongly imbalanced class problem with a long tail distribution as shown by the large difference between the maximum entropy and the actual entropy.
Previous works have already shown that objects are highly predictive of the predicates \cite{Zellers2017, Xu2017}. These observations highlight the relational nature of scene graphs and enforce the idea of building contextualized models. 

\section{Methodology} \label{sec:approach}

\begin{figure}[h] 
\centering
\def\svgwidth{\columnwidth}
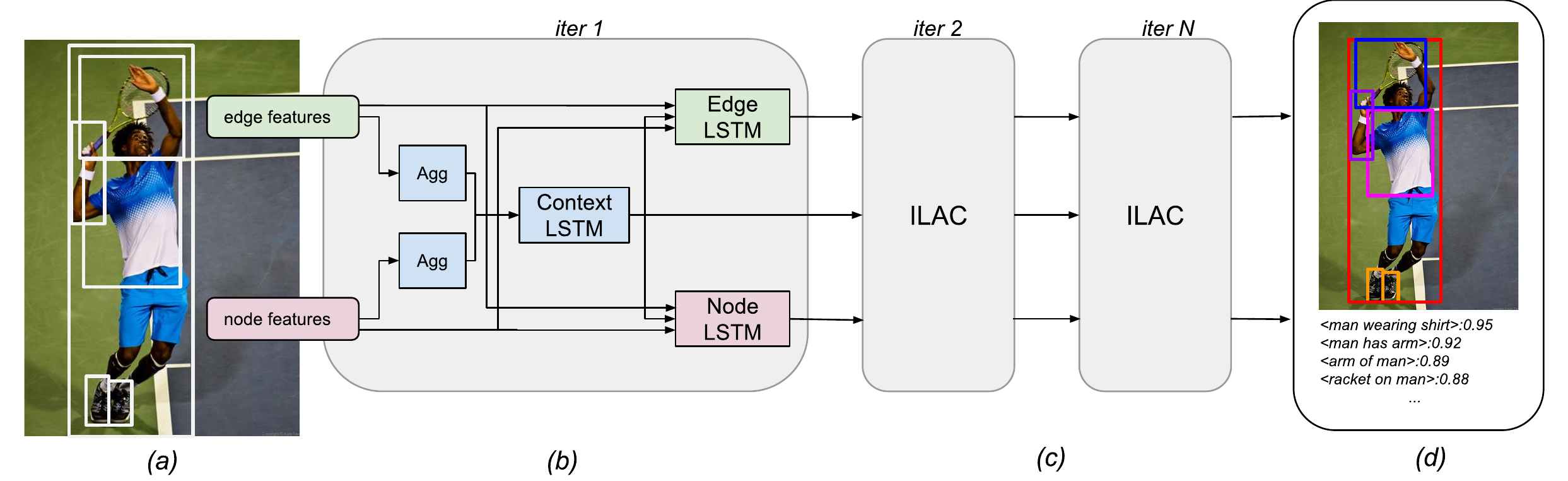
\caption{Pipeline overview. 
In (a), we start with bounding boxes (nodes) and union boxes (edges) associated with appearance features. In (b), we compute the image-level attentional context model (ILAC) from the node/edge aggregation (denoted by Agg). The context is used to refine each node and edge through the node and edge LSTM. In (c), we recursively apply this process until convergence. In (d), after $N$ steps, the object/predicate labels are derived with a classification layer. A confidence score is associated to each predicted phrase.
}
\label{fig:pipeline}
\end{figure}

ILAC takes as input an image with the bounding boxes of the objects, and outputs a scene graph. We build a graph $G=\{V,E,C\}$ represented by nodes, edges and a context. The nodes represent the objects, the edges represent the predicates and the context encodes image-level properties (e.g.~whether the scene captures a street or a room). 
We propose a composable directed graph-to-graph approach, where we iteratively refine the edge/node/context representations. The pipeline overview is shown in Figure~\ref{fig:pipeline}. 

%
\textbf{Initialization} The input features of the graph are a set of $N_o$ objects.
Each object $o_i$ is represented by its bounding box $b_i$, visual features $f_i \in \mathbb{R}^{4096}$ extracted from a pre-trained net and soft labels predicted by an object detector in isolation \footnote{We provide more details on the object detector in isolation in the supplemental.}, denoted by $\Tilde{p}(o_i) \in \mathbb{R}^{|Cls|}$.
We build a fully connected graph by connecting each object, resulting in $N_o(N_o-1)$ edges. The initial node $v_i \in \mathbb{R}^{d_v}$ and edge $e_{ij} \in \mathbb{R}^{d_e}$ representations are constructed as: 
\begin{align}
    v_i^{(0)} = W_o [\Tilde{p}(o_i); b_i; f_i] \quad \mbox{and} \quad  e_{ij}^{(0)} = W_r f_{ij}
\end{align}
where $[.;.]$ denotes the concatenation operation. $f_i$ and $f_{ij}$ are visual features from object $i$ and from the union box of objects $i$ and $j$ respectively. $W_o \in \mathbb{R}^{(d_v \times (|Cls|+|b_i|+|f_i|)}$ and $W_r \in \mathbb{R}^{(d_e \times (|f_{ij}|)}$  are learned embeddings. 
The image context $c^{(n)} \in \mathbb{R}^{d_c}$ is initialized with zeros. 

\textbf{Recursion}
We first infer the image context from the aggregation of the nodes and the edges using an attention mechanism \cite{Bahdanau2015, Velickovic2017}. The intuition behind is that some objects from the scene are more explanatory of the overall context than others. By computing a weighted sum of the objects, we allow the network to learn discriminative objects. For instance, to infer that we are in a \textit{street}, knowing that a \textit{traffic light} is in the scene is more explanatory than the presence of a \textit{man}. A similar reasoning applies to relations. The context gathers information from the objects $c_o^{(n)}$ and from the relations $c_{r}^{(n)}$. We model the object context $c_o^{(n)}$ as:
\begin{align} \label{eq:context_o}
    c_o^{(n+1)} = \sum_i^{N_o} \alpha_i^{(n)} v_i^{(n)} \quad \mbox{with} \quad \alpha_i^{(n)} = \frac{\exp(w_o^T \sigma(\Phi_o [v_i^{(n)} ; c^{(n)}]))}{\sum_j^{N_o} \exp(w_o^T \sigma(\Phi_o [v_j^{(n)} ; c^{(n)}]))}
\end{align}
where $\sigma$ is the $\tanh$ activation function, $\Phi_o \in \mathbb{R}^{d_{\Phi} \times (d_v + d_c)}$ and $w_o \in \mathbb{R}^{d_\Phi}$ are learned parameters. 
A similar attentional approach is used to model the relation context $c_r^{(n)}$. 
The node and edge context are then used to update the image context $c^{(n)}$ as:
\begin{align} \label{eq:context}
    c^{(n+1)} = f_c([c_o^{(n+1)} ; c_{r}^{(n+1)}], c^{(n)})
\end{align}
%
%
The edges are then updated as follows:
\begin{align}
    e_{ij}^{(n+1)} = f_e([v_i^{(n)} ; v_j^{(n)} ; c^{(n+1)}] , e_{ij}^{(n)})
\end{align}
where $c^{(n+1)}$ is the previously computed context update, $v_i^{(n)}$, $ v_j^{(n)}$ are the node of object $i$ and $j$, respectively. 
We then update the nodes as:
\begin{align}
    v_i^{(n+1)} = f_v([e_{i\bullet}^{(n)} ; c^{(n)}], v_i^{(n)})
\end{align}
where $e_{i\bullet}$ is the aggregation of all the incoming edges at node $i$ computed as in \eqref{eq:context_o}. 
Following previous works \cite{Palm2017, Sanchez-Gonzalez2018, Xu2017}, $f_c$, $f_e$ and $f_v$ are RNNs, namely LSTMs, where the cell hidden state represents the current edge/node/context. The hidden state is therefore iteratively refined as information from the graph diffuses. After $n$ iterations, the node and edge hidden states are passed through a classification layer to derive the labels. 
%
The hyperparameters and network dimensions are listed in Table \ref{tab:hyperparameters}. 
\begin{table}[h]
  \caption{Network dimensions and hyperparameters used.}
  \label{tab:hyperparameters}
  \centering
  \begin{tabular}{cccccccc}
  \toprule
    Name & $d_c$ & $d_v$ & $d_e$ & $d_{\Phi}$ & learning rate & batch size & iterations \\
    \midrule
    Value &   $512$  & $512$ & $512$ & $256$ & $10^{-4}$ & $8$  & $2$   \\
    \bottomrule
  \end{tabular}
\end{table}
%
\section{Experimental results} \label{sec:results}
\vspace{-1mm}

We used the Visual Genome dataset \cite{Krishna2017} cleansed by \cite{Xu2017} for training. It contains $150$ object categories and $50$ different predicates. During training, we use as loss function the sum of the cross-entropy loss of the objects and predicates. We evaluate the top-K recall (R@K) for the \textbf{predicate classification} (PredCls) and \textbf{scene graph classification} (SGCls). Note that we focus on the relational part of scene graphs and therefore we assume that the object location is known. 
We also report the contextualized object-detection accuracy
\footnote{We provide more details on the evaluation metrics in the supplemental section.}.

Experimental results are presented in Table \ref{table:result}. We compare our approach, ILAC, with IMP \cite{Xu2017}, neural motifs (NM) \cite{Zellers2017}, and a frequency baseline derived from statistics and object detection in isolation (FREQ) \cite{Zellers2017}.  
ILAC is found to be superior to IMP and FREQ, and have comparable performance with NM. However, ILAC uses $\approx 3 \times$ fewer parameters than NM, 
as can be seen in the last column of the table. This is achieved by keeping a unique global context and embedding the visual features into a lower-dimensional space.
More importantly, we improve the object detection in isolation (FREQ) by $1.5\%$, thus exhibiting the power of ILAC to extract the relational nature of scenes.
We also performed an ablation study to understand the contribution of each component. We first bypassed the image-level context module, bounding the model to node-to-edge and edge-to-node messages only (No-context in Table \ref{table:result}). The importance of the image-level context module is proven by experimental results as we observe a $3\%$ absolute performance gain for SGCls and $\approx 1\%$ for PredCls. This effect is more pronounced for SGCls than PredCls, as SGCls requires more relational reasoning and can therefore fully benefit from ILAC. Note that this observation matches the result of the entropy study. Beside having two iterations for ILAC, we also trained it with a single-pass graph network (1-iter). Having two iterations leads to an improvement of $\approx 0.6\%$, 
which shows the gain by iteratively refining the node/edge/context representations. It should be noted that we observed no further improvements upon adding additional iterations. 
\vspace{-2mm}
\begin{table}[h]
  \caption{Results expressed in $\%$. Number of parameters expressed in millions. All results were generated using the same object detector (from~\cite{Zellers2017}), evaluation service and test/train splitting to ensure fair comparisons. ILAC results have settings as listed in Table \ref{tab:hyperparameters}.}
  \label{table:result}
  \centering
  \begin{tabular}{lcccccccccc}
    \toprule
    \multicolumn{1}{c}{} &
    \multicolumn{2}{c}{PredCls} &
    \multicolumn{2}{c}{SGCls}  &
    \multicolumn{1}{c}{Object accuracy} &
    \multicolumn{1}{c}{Parameters}                   \\
    \cmidrule(r){2-3}
    \cmidrule(r){4-5}
    Model & R$@50$ & R$@100$ & R$@50$ & R$@100$     \\
    \midrule
    FREQ \cite{Zellers2017} & $75.2$ & $83.6$ & $39.0$ & $43.4$ & $66.1$ &  - \\
    IMP \cite{Xu2017} & $75.7$ & $82.9$  & $43.4$ & $47.2$ & $-$& $10.0$ \\
    NM \cite{Zellers2017} & $81.1$ & $88.3$ & $44.5$ & $47.7$ & $67.3$ & $40.0$ \\
    \textbf{ILAC} &  $\mathbf{80.4}$ & $\mathbf{88.4}$ & $\mathbf{43.7}$ & $\mathbf{47.6}$ & $\mathbf{67.6}$ & $\mathbf{14.8}$ \\
    \midrule
    No-context &  $78.8$ & $87.6$ & $40.3$ & $44.6$ & $66.2$ & $14.8$  \\
    ILAC (1-iter) &  $79.8$ & $87.9$ & $43.0$ & $47.0$  & $67.4$ & $14.8$\\
    \bottomrule
  \end{tabular}
\end{table}

\vspace{-4mm}
\section{Conclusion} \label{sec:conclusion}
\vspace{-1mm}
This paper introduces an efficient method for achieving scene graph generation, called the image-level attentional context (ILAC) modeling. The ILAC model learns to refine graph nodes and edges iteratively, then propagates contextual information across the graph. We demonstrate that sharing the information on a graph through a context agent allows to create competitive models that have less parameters than the state-of-the-art. 
More importantly, we show that image-level contexts can refine object detection by providing relational information.
Moreover, ILAC does not require domain-specific or external knowledge, which makes it easily pluggable on top of existing object detectors (with or without predicate information) to account for the relational nature of objects in images. 


\section*{Acknowledgements}

The authors would like to thank their IBM colleagues Anca-Nicoleta Ciubotaru, Andrea Giovannini, Antonio Foncubierta-Rodriguez and Apostolos Krystallidis for their feedback on this manuscript. 

\medskip
\small
\bibliographystyle{abbrv}
\bibliography{nips_2018}

\clearpage
\section*{Supplemental}



\subsection*{Dataset and implementation details}
As the original Visual Genome dataset is noisy, we used the pruned version provided by Xu et al.~\cite{Xu2017}. It contains $108,077$ annotated images, with an average of $6.2$ predicates and $11.5$ objects per image. $76631$ images were used for training, $5000$ for validation and $26446$ for testing.
The system was implemented using Pytorch 0.4.1 and trained on NVidia P100 GPUs. For each experiment, the system was trained for $15$ epochs using an Adam optimizer. The model that obtained best results with regard to validation was used for testing. The object detector in isolation is based on a Faster R-CNN backbone and was trained prior to ILAC. VGG16 pre-trained features were used for initialization. The object detector was then retrained for the specific $150$ object classes (plus background). To expose the network to negative relations (i.e.~objects not interacting), we randomly sampled during training object pairs without annotations such that each image contains $16$ predicates. 

\subsection*{Evaluation}
This section provides details about the evaluation metrics introduced in Section~\ref{sec:results}. The predicate classification (PredCls): given ground truth boxes and object labels, predict the predicates. Scene graph classification (SGCls): given ground truth boxes, predict the object labels and the predicates. We evaluate the recall for $k = {50, 100}$, meaning that we extract the top-K most likely phrases during testing and compare them with the annotations. 
We adopt the recall as the main metric because the annotations are often incomplete and subjective. The algorithm can suggest phrases that are visually correct but that were not annotated. The algorithms are evaluated in the unconstrained setting \cite{Newell2017}, meaning that we allow the same object pair to be described by several predicates. The object detection accuracy is computed as: given the image and the ground truth bounding boxes, predict the object label.  

\subsection*{Qualitative results}
\begin{figure*}[h]
    \centering
    \def\svgwidth{\columnwidth}
    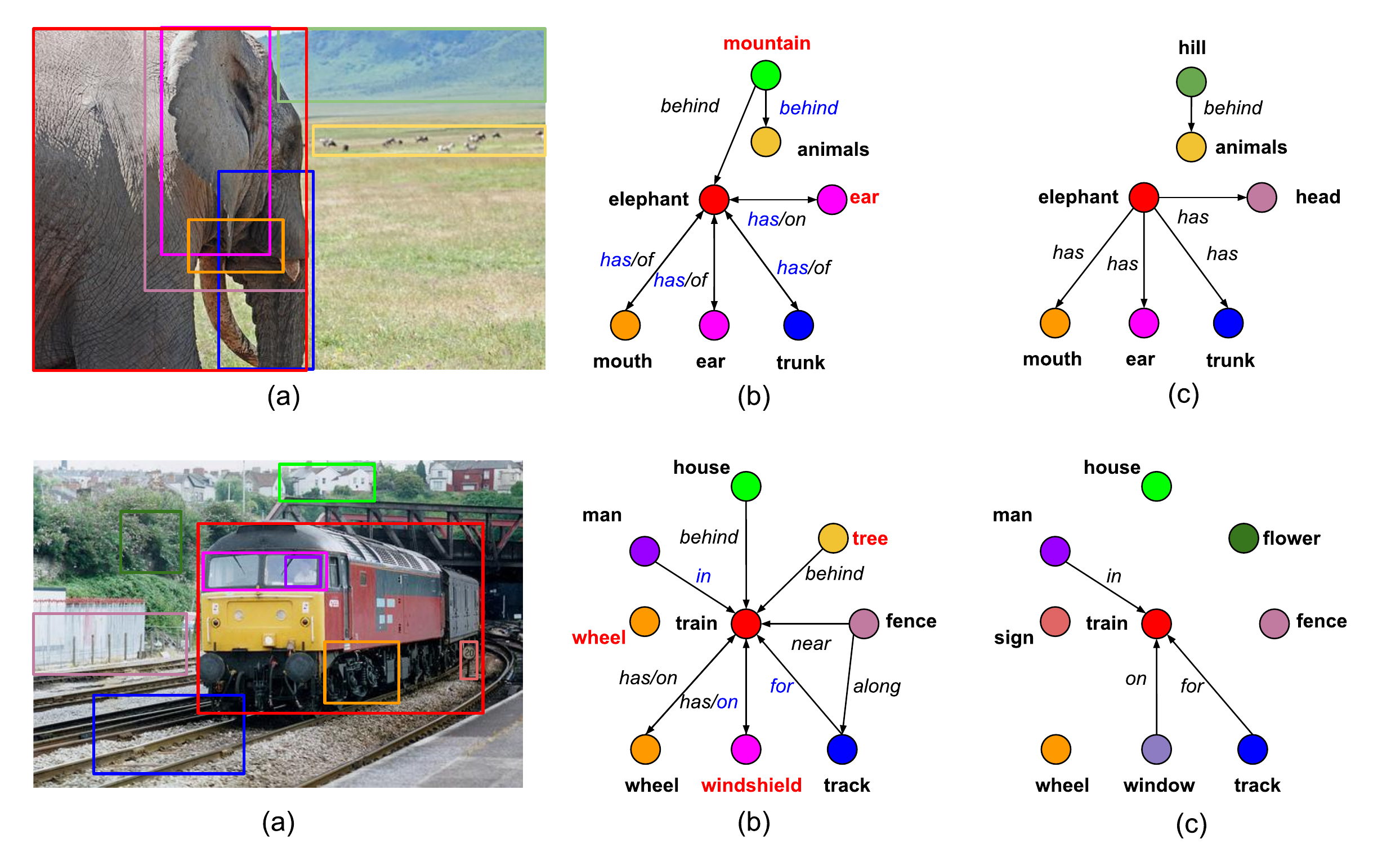
    \caption{Qualitative results from our ILAC model. (a) Original image with color-coded objects. (b) ILAC scene graph generation. Results obtained in the SGCls mode, by taking the predictions of the ten most likely phrases. Objects in red indicate a misclassification. Predicates in blue indicate a match with the ground truth. (c) Ground-truth scene graph. }
    \label{fig:qualitative_results}
\end{figure*}

Qualitative results are shown in Figure~\ref{fig:qualitative_results}. We note that among the top-10 predictions shown in (b), several are visually correct relations that do not appear in the ground truth. 

\end{document}